\documentclass[11pt,a4paper]{article}
\usepackage[hyperref]{acl2019}
\usepackage{times}
\usepackage{latexsym}
\usepackage{booktabs} 
\usepackage{multirow}
\usepackage{url}
\usepackage{amsmath}
\usepackage{amsfonts}
\usepackage{graphicx}
\usepackage{bm}
\usepackage{color}
\usepackage{diagbox}
\usepackage{xspace}
\makeatletter

\newcommand{\Rmnum}[1]{\expandafter\@slowromancap\romannumeral #1@}
\makeatother

\aclfinalcopy 

\title{DocRED: A Large-Scale Document-Level Relation Extraction Dataset}

\author{Yuan Yao$^{1}\thanks{\quad indicates equal contribution}$\hspace{0.5em}, Deming Ye$^{1*}$,  Peng Li$^{2}$, Xu Han$^{1}$, Yankai Lin$^{1}$, Zhenghao Liu$^{1}$, \\ \textbf{Zhiyuan Liu$^{1}$\thanks{\quad Corresponding author: Z.Liu(liuzy@tsinghua.edu.cn)}\hspace{0.5em}, Lixin Huang$^{1}$, Jie Zhou$^{2}$, Maosong Sun$^{1}$}\\
$^{1}$Department of Computer Science and Technology, Tsinghua University, Beijing, China\\
Institute for Artificial Intelligence, Tsinghua University, Beijing, China\\
State Key Lab on Intelligent Technology and Systems, Tsinghua University, Beijing, China \\
$^{2}$Pattern Recognition Center, WeChat AI, Tencent Inc.\\
\texttt{\{yuan-yao18,ydm18\}@mails.tsinghua.edu.cn}
}

\date{}

\begin{document}
\maketitle
\begin{abstract}
Multiple entities in a document generally exhibit complex inter-sentence relations, and cannot be well handled by existing relation extraction (RE) methods that typically focus on extracting intra-sentence relations for single entity pairs. In order to accelerate the research on document-level RE, we introduce DocRED, a new dataset constructed from Wikipedia and Wikidata with three features: (1) DocRED annotates both named entities and relations, and is the largest human-annotated dataset for document-level RE from plain text; (2) DocRED requires reading multiple sentences in a document to extract entities and infer their relations by synthesizing all information of the document; (3) along with the human-annotated data, we also offer large-scale distantly supervised data, which enables DocRED to be adopted for both supervised and weakly supervised scenarios. In order to verify the challenges of document-level RE, we implement recent state-of-the-art methods for RE and conduct a thorough evaluation of these methods on DocRED. Empirical results show that DocRED is challenging for existing RE methods, which indicates that document-level RE remains an open problem and requires further efforts. Based on the detailed analysis on the experiments, we discuss multiple promising directions for future research. We make DocRED and the code for our baselines publicly available at \url{https://github.com/thunlp/DocRED}.
\end{abstract}

\section{Introduction}
\begin{figure}[t]
    \centering
    \includegraphics[width=\columnwidth]{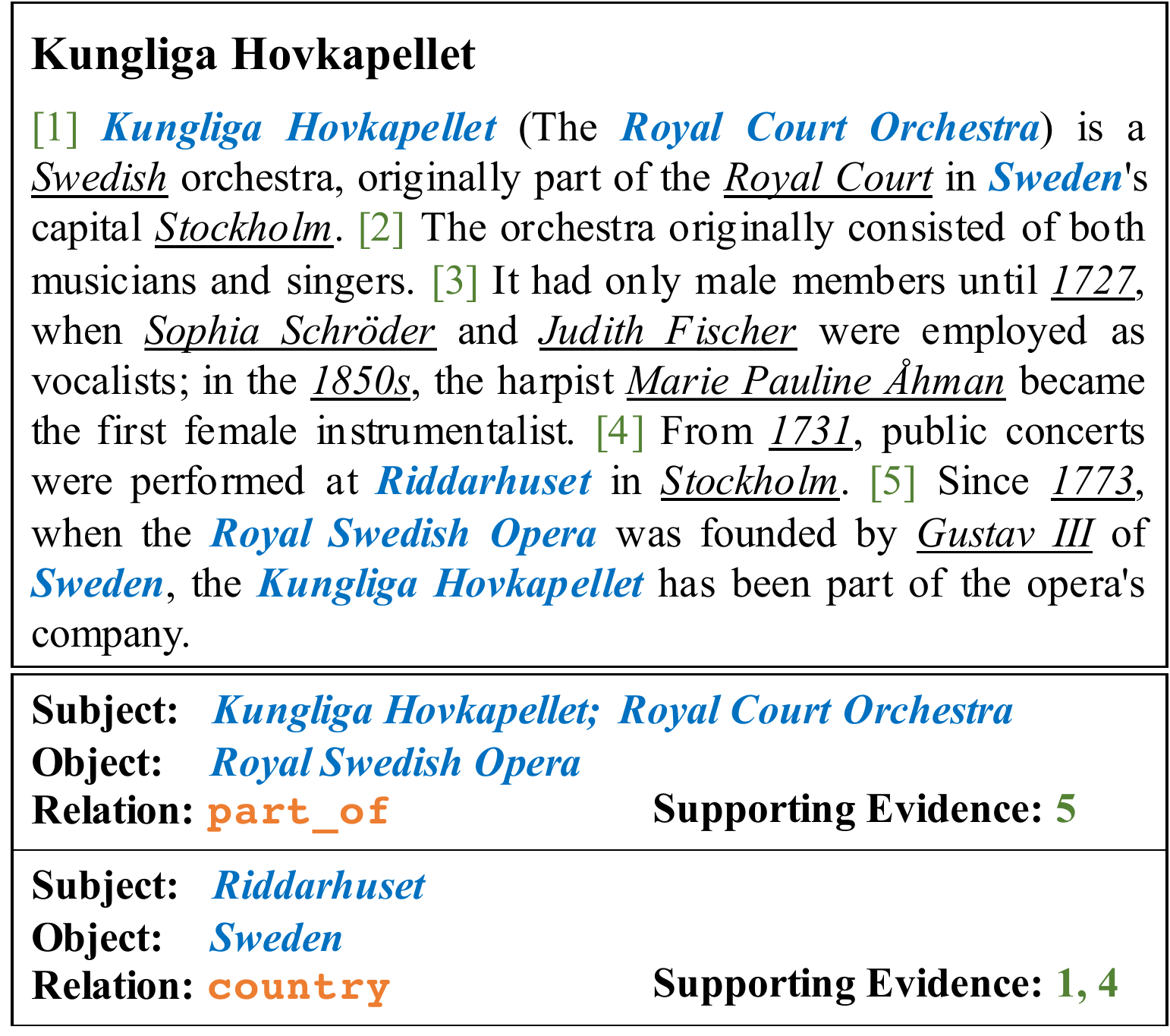}
    \caption{An example from DocRED. Each document in DocRED is annotated with named entity mentions, coreference information, intra- and inter-sentence relations, and supporting evidence. $2$ out of the $19$ relation instances annotated for this example document are presented, with  named entity mentions involved in these instances colored in blue and other named entity mentions underlined for clarity. Note that mentions of the same subject (e.g., \emph{Kungliga Hovkapellet} and \emph{Royal Court Orchestra}) are identified as shown in the first relation instance.}
    \label{fig:example}
    \vspace{-1.5em}
\end{figure}

The task of relation extraction (RE) is to identify relational facts between entities from plain text, which plays an important role in large-scale knowledge graph construction. Most existing RE work focuses on sentence-level RE, i.e., extracting relational facts from a single sentence. In recent years, various neural models have been explored to encode relational patterns of entities for sentence-level RE, and achieve state-of-the-art performance~\cite{socher2012semantic,DBLP:conf/coling/ZengLLZZ14,zeng2015distant,santos2015classifying,xiao2016semantic,cai2016bidirectional,lin2016neural,wu2017adversarial,qin2018robust,han2018hierarchical}.

Despite these successful efforts, sentence-level RE suffers from an inevitable restriction in practice: a large number of relational facts are expressed in multiple sentences. Taking Figure~\ref{fig:example} as an example, multiple entities are mentioned in the document and exhibit complex interactions. In order to identify the relational fact (\textit{Riddarhuset}, \texttt{country}, \textit{Sweden}), one has to first identify the fact that \textit{Riddarhuset} is located in \textit{Stockholm} from Sentence 4, then identify the facts \textit{Stockholm} is the capital of \textit{Sweden} and \textit{Sweden} is a country from Sentence 1, and finally infer from these facts that the sovereign state of \textit{Riddarhuset} is \textit{Sweden}. The process requires reading and reasoning over multiple sentences in a document, which is intuitively beyond the reach of sentence-level RE methods. According to the statistics on our human-annotated corpus sampled from Wikipedia
documents, at least $40.7\%$ relational facts can only be extracted from multiple sentences, which is not negligible. \newcite{swampillai2010MUC} and \newcite{verga2018simultaneously} have also reported similar observations. Therefore, it is necessary to move RE forward from sentence level to document level.

The research on document-level RE requires a large-scale annotated dataset for both training and evaluation. Currently, there are only a few datasets for document-level RE. \newcite{quirk2017distant} and \newcite{peng2017nary} build two distantly supervised datasets without human annotation, which may make the evaluation less reliable. BC5CDR~\cite{li2016biocreative} is a human-annotated document-level RE dataset consisting of $1,500$ PubMed documents, which is in the specific domain of biomedicine considering only the ``chemical-induced disease'' relation, making it unsuitable for developing general-purpose methods for document-level RE. \newcite{levy2017re-as-rc} extract relational facts from documents by answering questions using reading comprehension methods, where the questions are converted from entity-relation pairs. As the dataset proposed in this work is tailored to the specific approach, it is also unsuitable for other potential approaches for document-level RE. In summary, existing datasets for document-level RE either only  have a small number of manually-annotated relations and entities, or exhibit noisy annotations from distant supervision, or serve specific domains or approaches. In order to accelerate the research on document-level RE, we urgently need a large-scale, manually-annotated, and general-purpose document-level RE dataset.

In this paper, we present DocRED, a large-scale human-annotated document-level RE dataset constructed from Wikipedia and Wikidata~\cite{DBLP:conf/semweb/ErxlebenGKMV14,vrandevcic2014wikidata}.
DocRED is constructed with the following three features: (1) DocRED contains $132,375$ entities and $56,354$ relational facts annotated on $5,053$ Wikipedia documents, making it the largest human-annotated document-level RE dataset. (2) As at least $40.7\%$ of the relational facts in DocRED can only be extracted from multiple sentences, DocRED requires reading multiple sentences in a document to recognize entities and inferring their relations by synthesizing all information of the document. This distinguishes DocRED from those sentence-level RE datasets. (3) We also provide large-scale distantly supervised data to support weakly supervised RE research.

To assess the challenges of DocRED, we implement recent state-of-the-art RE methods and conduct thorough experiments on DocRED under various settings. Experimental results show that the performance of existing methods declines significantly on DocRED, indicating the task document-level RE is more challenging than sentence-level RE and remains an open problem. Furthermore, detailed analysis on the results also reveals multiple promising directions worth pursuing. 

\section{Data Collection}
Our ultimate goal is to construct a dataset for document-level RE from plain text, which requires necessary information including named entity mentions, entity coreferences, and relations of all entity pairs in the document. To facilitate more RE settings, we also provide supporting evidence information for relation instances. In the following sections, we first introduce the collection process of the human-annotated data, and then describe the process of creating the large-scale distantly supervised data.

\subsection{Human-Annotated Data Collection}
\label{Sec:Human-Annotated Data Collection}

Our human-annotated data is collected in four stages: (1) Generating distantly supervised annotation for Wikipedia documents. (2) Annotating all named entity mentions in the documents and coreference information. (3) Linking named entity mentions to Wikidata items. (4) Labeling relations and corresponding supporting evidence. 

Following ACE annotation process~\cite{doddington2004automatic}, both Stage 2 and 4 require three iterative passes over the data: (1) Generating named entity using named entity recognition (NER) models, or relation recommendations using distant supervision and RE models. (2) Manually correcting and supplementing recommendations. (3) Reviewing and further modifying the annotation results from the second pass for better accuracy and consistency. To ensure the annotators are well trained, a principled training procedure is adopted and the annotators are required to pass test tasks before annotating the dataset. And only carefully selected experienced annotators are qualified for the third pass annotation.

To provide a strong alignment between text and KBs, our dataset is constructed from the complete English Wikipedia document collection and Wikidata~\footnote{We use the 2018-5-24 dump of English Wikipedia and 2018-3-20 dump of Wikidata.}, which is a large-scale KB tightly integrated with Wikipedia. We use the introductory sections from Wikipedia documents as the corpus, as they are usually high-quality and contain most of the key information. 

\smallskip
\noindent
\textbf{Stage 1: Distantly Supervised Annotation Generation.}
\label{Sec: distant}
To select documents for human annotation, we align Wikipedia documents with Wikidata under the distant supervision assumption~\cite{mintz2009distant}. Specifically, we first perform named entity recognition using {spaCy}\footnote{https://spacy.io}. Then these named entity mentions are linked to Wikidata items, where named entity mentions with identical KB IDs are merged. Finally, relations between each merged named entity pair in the document are labeled by querying Wikidata. Documents containing fewer than $128$ words are discarded. To encourage reasoning, we further discard documents containing fewer than $4$ entities or fewer than $4$ relation instances, resulting in $107,050$ documents with distantly supervised labels, where we randomly select $5,053$ documents and the most frequent $96$ relations for human annotation. 
 
\smallskip
\noindent
\textbf{Stage 2: Named Entity and Coreference Annotation.} 
Extracting relations from document requires first recognizing named entity mentions and identifying mentions referring to the same entities within the document. To provide high-quality named entity mentions and coreference information, we ask human annotators first to review, correct and supplement the named entity mention recommendations generated in Stage 1, and then merge those different mentions referring to the same entities, which provides extra coreference information. The resulting intermediate corpus contains a variety of named entity types including person, location, organization, time, number and names of miscellaneous entities that do not belong to the aforementioned types. 

\smallskip
\noindent
\textbf{Stage 3: Entity Linking.}
 In this stage, we link each named entity mention to multiple Wikidata items to provide relation recommendations from distant supervision for the next stage. To be specific, each named entity mention is associated with a Wikidata item candidate set~\footnote{To avoid losing relation recommendations due to prediction errors in entity linking, we include multiple linking results from different approaches in the candidate set.} consisting of all Wikidata items whose names or aliases literally match it. We further extend the candidate set using Wikidata items hyperlinked to the named entity mention by the document authors, and recommendations from an entity linking toolkit TagMe~\cite{Tagme}.  Specially, numbers and time are semantically matched.

\smallskip
\noindent
\textbf{Stage 4: Relation and Supporting Evidence Collection.}
\label{Sec:Stage 4}
The annotation of relation and supporting evidence is based on the named entity mentions and coreference information in Stage 2, and faces two main challenges. The first challenge comes from the large number of potential entity pairs in the document. On the one hand, given the quadratic number of potential entity pairs with regard to entity number ($19.5$ entities on average) in a document, exhaustively labeling relations between each entity pair would lead to intensive workload. On the other hand, most entity pairs in a document do not contain relations. The second challenge lies in the large number of fine-grained relation types in our dataset. Thus it is not feasible for annotators to label relations from scratch.

We address the problem by providing human annotators with recommendations from RE models, and distant supervision based on entity linking (Stage 3). On average, we recommend $19.9$ relation instances per document from entity linking, and $7.8$ from RE models for supplement. We ask the annotators to review the recommendations, remove the incorrect relation instances and  supplement omitted ones. We also ask the annotators to further select all sentences that support the reserved relation instances as supporting evidence. Relations reserved must be reflected in the document, without relying on external world knowledge. Finally $57.2\%$ relation instances from entity linking and $48.2\%$  from RE models are reserved.

\begin{table*}
    \centering
    \small
    \begin{tabular}{lrrrrrrr}
    \toprule
    Dataset & \# Doc. & \# Word & \# Sent. & \# Ent. & \# Rel. & \# Inst. &\# Fact \\
    \midrule
    
    SemEval-2010 Task 8 & - & 205k &10,717& 21,434 & 9 & 8,853 & 8,383\\
    ACE 2003-2004 & - & 297k & 12,783& 46,108 & 24 & 16,771& 16,536\\
    TACRED & - & 1,823k&53,791 & 152,527 & 41 &21,773& 5,976\\
    FewRel& - & 1,397k & 56,109 & 72,124 & 100 & 70,000 & 55,803\\
    BC5CDR & 1,500 & 282k &11,089 & 29,271 & 1 & 3,116 & 2,434\\
    DocRED (Human-annotated) & 5,053& 1,002k & 40,276 & 132,375& 96 & 63,427 & 56,354\\
    DocRED (Distantly Supervised) & 101,873& 21,368k & 828,115& 2,558,350& 96 &1,508,320 &881,298\\
    \bottomrule
    \end{tabular}
    \caption{Statistics of RE datasets (Doc.: document, Sent.: sentence, Ent.: entity, Rel.: relation type, Inst.: relation instance, Fact: relational fact). The first four are sentence-level RE datasets.}
    \vspace{-0.5em}
    \label{Table:data_statistics}
\end{table*}

\subsection{Distantly Supervised Data Collection}
\label{Sec:Distantly Supervised Data Collection}
In addition to the human-annotated data, we also collect large-scale distantly supervised data to promote weakly supervised RE scenarios. We remove the $5,053$ human-annotated documents from the $106,926$ documents, and use the rest $101,873$ documents as the corpus of distantly supervised data. To ensure that the distantly supervised data and human-annotated data share the same entity distribution, named entity mentions are re-identified using Bidirectional Encoder Representations from Transformers (BERT)~\cite{devlin2018bert} that is fine-tuned  on the human-annotated data collected in Sec.~\ref{Sec:Human-Annotated Data Collection} and achieves $90.5\%$ F1 score. We link each named entity mention to one Wikidata item by a heuristic-based method, which jointly considers the frequency of a target Wikidata item and its relevance to the current document. Then we merge the named entity mentions with identical KB IDs. Finally, relations between each merged entity pair are labeled via distant supervision.

\section{Data Analysis}
\label{Sec:Data Analysis}
In this section, we analyze various aspects of DocRED to provide a deeper understanding of the dataset and the task of document-level RE.

\smallskip
\noindent
\textbf{Data Size.} Table~\ref{Table:data_statistics} shows statistics of DocRED and some representative RE datasets, including sentence-level RE datasets SemEval-2010 Task 8~\cite{hendrickx2009semeval}, ACE 2003-2004~\cite{doddington2004automatic}, TACRED~\cite{zhang2017position}, FewRel~\cite{han2018fewrel} and document-level RE dataset BC5CDR~\cite{li2016biocreative}. We find that DocRED is larger than existing datasets in many aspects, including the number of documents, words, sentences, entities, especially in aspects of relation types, relation instances and relational facts. We hope the large-scale DocRED dataset could drive relation extraction from sentence level forward to document level.

\smallskip
\noindent
\textbf{Named Entity Types.}
DocRED covers a variety of entity types, including person ($18.5\%$), location ($30.9\%$), organization ($14.4\%$), time ($15.8\%$) and number ($5.1\%$). It also covers  a diverse set of miscellaneous entity names ($15.2\%$) not belonging to the aforementioned types, such as events, artistic works and laws. Each entity is annotated with $1.34$ mentions on average.

\smallskip
\noindent
\textbf{Relation Types.}
Our dataset includes $96$ frequent relation types from Wikidata. A notable property of our dataset is that the relation types cover a broad range of categories, including relations relevant to science ($33.3\%$), art ($11.5\%$), time ($8.3\%$), personal life ($4.2\%$), etc., which means the relational facts are not constrained in any specific domain. In addition, the relation types are organized in a well-defined hierarchy and taxonomy, which could provide rich information for document-level RE systems.  %

\newcommand{\tabincell}[2]{\begin{tabular}{@{}#1@{}}#2\end{tabular}}
    \definecolor{blue_h}{RGB}{68,114,196}
    \definecolor{olive}{RGB}{84,130,53}
    \definecolor{rel}{RGB}{237,125,49}
    \begin{table*}[htbp]
    \begin{center}
    \small
    \begin{tabular}{p{0.15\textwidth}  r p{0.69\textwidth}}
    \toprule
    Reasoning Types & \% & Examples\\
    \midrule
    Pattern recognition & 38.9 &  $\!${\color{olive} [1]} {\color{blue} \textit{\textbf{Me Musical Nephews}}} is a {\color{blue_h} \textbf{1942}} one-reel animated cartoon directed by Seymour Kneitel and animated by Tom Johnson and George Germanetti. {\color{olive}[2] }Jack Mercer and Jack Ward wrote the script. ... \\
    & & \textbf{Relation:} {\color{rel} \textbf{\texttt{publication\_date}}} \ \ \, \; \qquad \qquad  \textbf{Supporting Evidence:} {\color{olive} \textbf{1}} \\
    
    \midrule
    Logical reasoning & 26.6 & $\!${\color{olive}[1] } ``Nisei" is the ninth episode of the third season of the American science fiction television series The X-Files. ... {\color{olive}[3] }It was directed by David Nutter, and written by {\color{blue_h} \textbf{Chris Carter}}, Frank Spotnitz and Howard Gordon. ... {\color{olive}[8] }The show centers on FBI special agents {\color{blue} \textbf{\textit{Fox Mulder}}} (David Duchovny) and Dana Scully (Gillian Anderson) who work on cases linked to the paranormal, called X-Files.  ... \\
    & & \textbf{Relation:} {\color{rel} \textbf{\texttt{creator}}} \qquad \qquad \qquad \qquad \qquad \ \ \textbf{Supporting Evidence:} {\color{olive} \textbf{1, 3, 8}} \\   
    \midrule
    Coreference reasoning & 17.6 &$\!${\color{olive}[1] }{\color{blue} \textbf{\textit{Dwight Tillery}}} is an American politician of the Democratic Party who is active in local politics of Cincinnati, Ohio. ... {\color{olive}[3] }He also holds a law degree from the {\color{blue_h} \textbf{University of Michigan Law School}}. {\color{olive}[4] }{\color{blue} \textbf{\textit{Tillery}}} served as mayor of Cincinnati from 1991 to 1993. \\
    & & \textbf{Relation:} {\color{rel} \texttt{\textbf{educated\_at}}} \ \qquad \qquad \qquad \qquad  \textbf{Supporting Evidence:} {\color{olive} \textbf{1, 3}} \\   
    \midrule
    Common-sense reasoning &16.6  &$\!${\color{olive} [1]} {\color{blue} \textbf{\textit{William Busac}}} (1020-1076), son of William I, Count of Eu, and his wife Lesceline. ... {\color{olive} [4]} {\color{blue} \textbf{\textit{William}}} appealed to King Henry I of France, who gave him in marriage {\color{blue_h}\textbf{Adelaide}}, the heiress of the county of Soissons. {\color{olive} [5]} {\color{blue_h} \textbf{Adelaide}} was daughter of Renaud I, Count of Soissons, and Grand Master of the Hotel de France. ... {\color{olive} [7]} {\color{blue} \textbf{\textit{William}}} and {\color{blue_h} \textbf{Adelaide}} had four children: ... 
    \\
    & & \textbf{Relation:} {\color{rel} \texttt{\textbf{spouse}}} \qquad \qquad \qquad \qquad \qquad \quad   \textbf{Supporting Evidence:} {\color{olive} \textbf{4, 7}} \\     
    
    \bottomrule
    \end{tabular}
    \end{center}
    \caption{Types of reasoning required for document-level RE on DocRED. The rest $0.3\%$ requires other types of reasoning, such as temporal reasoning. The {\color{blue}\bf \textit{head}}, {\color{blue_h} \bf tail} and {\color{rel} \bf \texttt{relation}} are colored accordingly.}
    \label{Table:reasoning types}
\end{table*}

\smallskip
\noindent
\textbf{Reasoning Types.}
We randomly sampled $300$ documents from dev and test set, which contain $3,820$ relation instances, and manually analyze the reasoning types required to extract these relations.  Table \ref{Table:reasoning types} shows statistics of major reasoning types in our dataset. From the statistics on reasoning types, we have the following observations:
(1) Most of the relation instances ($61.1\%$) require reasoning to be identified, and only $38.9\%$ relation instances can be extracted via simple pattern recognition, which indicates that reasoning is essential for document-level RE. 
(2) In relation instances with reasoning, a majority ($26.6\%$) require logical reasoning, where the relations between two entities in question are indirectly established by a bridge entity. Logical reasoning requires RE systems to be capable of modeling interactions between multiple entities.
(3) A notable number of relation instances ($17.6\%$) need coreference reasoning, where coreference resolution must be performed first to identify target entities in a rich context. 
(4) A similar proportion of relation instances ($16.6\%$) has to be identified based on common-sense reasoning, where readers need to combine relational facts from the document with common-sense to complete the relation identification. In summary, DocRED requires rich reasoning skills for synthesizing all information of the document.


\smallskip
\noindent
\textbf{Inter-Sentence Relation Instances.} We find that each relation instance is associated with $1.6$ supporting sentences on average, where $46.4\%$ relation instances are associated with more than one supporting sentence. 
Moreover, detailed analysis reveals that $40.7\%$ relational facts can only be extracted from multiple sentences, indicating that DocRED is a good benchmark for document-level RE. We can also conclude that the abilities of reading, synthesizing and reasoning over multiple sentence are essential for document-level RE.


\section{Benchmark Settings}
\label{Sec:Benchmark}
We design two benchmark settings for supervised and weakly supervised scenarios respectively. For both settings, RE systems are evaluated on the high-quality human-annotated dataset, which provides more reliable evaluation results for document-level RE systems. The statistics of data used for the two settings are shown in Table~\ref{tab:data split}.

\begin{table}
    \begin{center}
    \small
    \begin{tabular}{p{2em}p{1em}rrrr}
    \toprule
    \multicolumn{2}{c}{Setting} & \# Doc. & \# Rel. & \# Inst. &\# Fact \\
    \midrule
    Train & W & 101,873 & 96& 1,508,320 & 881,298 \\
     & S & 3,053 & 96& 38,269 & 34,715 \\
    Dev & S,W & 1,000  & 96& 12,332 & 11,790 \\
    Test & S,W & 1,000 & 96& 12,842 & 12,101 \\
    \bottomrule
    \end{tabular}
    \end{center}
    \caption{Statistics of data used for the two benchmark settings (Sec.~\ref{Sec:Benchmark}): supervised setting (S) and weakly supervised setting (W).}
    \label{tab:data split}
    \vspace{-0.5em}
\end{table}

\smallskip
\noindent
\textbf{Supervised Setting.}
In this setting, only human-annotated data is used, which are randomly split into training, development and test sets. The supervised setting brings up two challenges for document-level RE systems as follows:

The first challenge comes from the rich reasoning skills required for performing document-level RE. As shown in Sec.~\ref{Sec:Data Analysis}, about $61.1\%$ relation instances depend on complex reasoning skills other than pattern recognition to be extracted, which requires RE systems to step beyond recognizing simple patterns in a single sentence, and reason over global and complex information in a document.

The second challenge lies in the high computational cost of modeling long documents and the massive amount of potential entity pairs in a document, which is quadratic with regard to entity number ($19.5$ entities on average) in a document. As a result, RE systems that model context information with algorithms of quadratic or even higher computational complexity such as~\cite{DBLP:conf/emnlp/SorokinG17,christopoulou2019walk} are not efficient enough for document-level RE. Thus the efficiency of context-aware RE systems needs to be further improved to be applicable in document-level RE.

\smallskip
\noindent
\textbf{Weakly Supervised Setting.}
This setting is identical to the supervised setting except that the training set is replaced with the distantly supervised data (Sec.~\ref{Sec:Distantly Supervised Data Collection}). In addition to the aforementioned two challenges, the inevitable wrong labeling problem accompanied with distantly supervised data is a major challenge for RE models under weakly supervised setting. Many efforts have been devoted to alleviating the wrong labeling problem in sentence-level RE~\cite{DBLP:conf/pkdd/RiedelYM10,hoffmann2011knowledge,surdeanu2012multi,lin2016neural}. However, noise in document-level distantly supervised data is significantly more than its counterpart in sentence-level. For example, for the recommended relation instances whose head and tail entities co-occur in the same sentence (i.e. intra-sentence relation instance) in Stage 4 of human-annotated data collection (Sec.~\ref{Sec:Stage 4}), $41.4\%$ are labeled as incorrect, while $61.8\%$ inter-sentence relation instances are labeled as incorrect, indicating the wrong labeling problem is more challenging for weakly supervised document-level RE.
Therefore, we believe offering distantly supervised data in DocRED will accelerate the development of distantly supervised methods for document-level RE. Moreover, it is also possible to jointly leverage distantly supervised data and human-annotated data to further improve the performance of RE systems.

\section{Experiments}
To assess the challenges of DocRED, we conduct comprehensive experiments to evaluate state-of-the-art RE systems on the dataset. Specifically, we conduct experiments under both supervised and weakly supervised benchmark settings. We also assess human performance and analyze the performance for different supporting evidence types. In addition, we conduct ablation study to investigate the contribution of different features. Through detailed analysis, we discuss several future directions for document-level RE.


\begin{table*}
    \begin{center}
    \small
    \begin{tabular}{l|cccc|cccc}
    \toprule
    \multirow{2}{*}{Model} & \multicolumn{4}{c|}{Dev}& \multicolumn{4}{c}{Test} \\
     & Ign\,F1 & Ign\,AUC & F1 & AUC & Ign\,F1 & Ign\,AUC & F1 & AUC \\
    \midrule
    \multicolumn{9}{c}{Supervised Setting}\\
    \midrule
    CNN & 41.58 &  36.85 & 43.45 & 39.39 & 40.33 & 36.24& 42.26 & 38.91\\
    LSTM & 48.44 & 46.62 &50.68 & 49.48 & 47.71 & 46.27 & 50.07 & 49.25\\
    BiLSTM & {48.87} & \textbf{47.61} & 50.94 & \textbf{50.26}  &\textbf{48.78} & \textbf{47.61} & \textbf{51.06} & \textbf{50.43}\\
    Context-Aware & \textbf{48.94} & {47.22}& \textbf{51.09}& {50.17}& {48.40} & {46.54} & {50.70} & {49.64}\\    
    \midrule
    
    \multicolumn{9}{c}{Weakly Supervised Setting}\\
    \midrule
    CNN & 33.24 & 23.17 & 42.76 & 37.99 & 32.33 & 21.83 & 42.00& 36.84\\
    LSTM & 39.37& 22.39 & 49.92 & 42.79 & 38.27 & 21.74 & 48.88 & 41.35\\
    BiLSTM & \textbf{41.44} & \textbf{23.21} & \textbf{51.72} & \textbf{44.44}  & 39.15 & \textbf{22.14} & 49.80 & \textbf{42.87}\\
    Context-Aware & {40.47} & 22.56 & {51.39}& 43.00 & \textbf{39.16} & 21.58 & \textbf{50.12} & {41.51}\\
    \bottomrule
    \end{tabular}
    \end{center}
    \caption{Performance of different RE models on DocRED (\%).}
    \label{table:main results}
\end{table*}

\smallskip
\noindent
\textbf{Models.}
We adapt four state-of-the-art RE models to document-level RE scenario, including a CNN~\cite{DBLP:conf/coling/ZengLLZZ14} based model, an LSTM~\cite{DBLP:journals/neco/HochreiterS97} based model, a  bidirectional LSTM (BiLSTM)~\cite{cai2016bidirectional} based model and the Context-Aware model~\cite{DBLP:conf/emnlp/SorokinG17} originally designed for leveraging contextual relations to improve intra-sentence RE.
The first three models differ only at the encoder used for encoding the document and will be explained in detail in the rest of this section. We refer the readers to the original paper for the details of the Context-Aware model for space limitation.

The CNN/LSTM/BiLSTM based models first encode a document $\mathcal{D}=\{w_i\}_{i=1}^n$ consisting of $n$ words into a hidden state vector sequence $\{\bm{h}_i\}_{i=1}^n$ with CNN/LSTM/BiLSTM as encoder, then compute the representations for entities, and finally predict relations for each entity pair.

For each word,  the features fed to the encoder is the concatenation of its GloVe word embedding~\cite{DBLP:conf/emnlp/PenningtonSM14}, entity type embedding and coreference embedding. The entity type embedding is obtained by mapping the entity type (e.g., PER, LOC, ORG) assigned to the word into a vector using an embedding matrix. The entity type is assigned by human for the human-annotated data, and by a fine-tuned BERT model for the distantly supervised data. 
Named entity mentions corresponding to the same entity are assigned with the same entity id, which is determined by the order of its first appearance in the document. And the entity ids are mapped into vectors as the coreference embeddings.

For each named entity mention $m_k$ ranging from the $s$-th word to the $t$-th word, we define its representation as $\bm{m}_k =\frac{1}{t-s+1}\sum_{j=s}^t\bm{h}_j$. And the representation of an entity $e_i$ with $K$ mentions is computed as the average of the representations of these mentions: $\bm{e}_i = \frac{1}{K} \sum_k\bm{m}_k$.

We treat relation prediction as a multi-label classification problem. Specially, for each entity pair $(e_i, e_j)$, we first concatenate the entity representations with relative distance embeddings, and then use a bilinear function to compute the probability for each relation type:
\begin{gather}
\hat{\bm{e}}_i=[\bm{e}_i;\bm{E}(d_{ij})],\;\hat{\bm{e}}_j=[\bm{e}_j;\bm{E}(d_{ji})]\label{eqn:distance}\\
P(r|e_i, e_j)=\mathrm{sigmoid}(\hat{\bm{e}}^{\mathrm{T}}_i \bm{W}_r\hat{\bm{e}_j} + b_r)
\end{gather}
where $[\cdot;\cdot]$ denotes concatenation, $d_{ij}$ and $d_{ji}$ are the relative distances of the first mentions of the two entities in the document, $\bm{E}$ is an embedding matrix,  $r$ is a relation type, and $\bm{W}_r$, $b_r$ are relation type dependent trainable parameters.

\smallskip
\noindent
\textbf{Evaluation Metrics.}
Two widely used metrics F1 and AUC are used in our experiments. However, some relational facts present in both the training and dev/test sets, thus a model may memorize their relations during training and achieve a better performance on the dev/test set in an undesirable way, introducing evaluation bias. However, the overlap in relational facts between the training and dev/test sets is inevitable, since many common relational facts are likely to be shared in different documents. Therefore, we also report the F1 and AUC scores excluding those relational facts shared by the training and dev/test sets, denoted as Ign\,F1 and Ign\,AUC, respectively. 

\smallskip
\noindent
\textbf{Model Performance.}
Table~\ref{table:main results} shows the experimental results under the supervised and weakly supervised settings, from which we have the following observations:
(1) Models trained with human-annotated data generally outperform their counterparts trained on distantly supervised data. This is because although large-scale distantly supervised data can be easily obtained via distant supervision, the wrong-labeling problem may harm the performance of RE systems, which makes weakly supervised setting a more difficult scenario. 
(2) An interesting exception is that LSTM, BiLSTM and Context-Aware trained on distantly supervised data achieve comparable F1 scores as those trained on human-annotated data but significantly lower scores on the other metrics, indicating that the overlap entity pairs between training and dev/test sets indeed cause evaluation biases. Therefore, reporting Ign\,F1 and Ign\,AUC is necessary.
(3) Models leveraging rich contextual information generally achieve better performances. LSTM and BiLSTM outperform CNN, indicating the effectiveness of modeling long-dependency semantics in document-level RE. Context-Aware achieves competitive performance, however, it cannot significantly outperform other neural models. It indicates that it is beneficial to consider the association of multiple relations in document-level RE, whereas the current models are not capable of utilizing inter-relation information well.

\begin{table}
\centering
\small

\begin{tabular}{l|c c c| c c c}
\toprule

\multirow{2}{*}{Method} & \multicolumn{3}{c|}{RE}  & \multicolumn{3}{c}{RE+Sup} \\
 & P & R & F1 & P & R & F1\\
 
\midrule

 Model  & 55.6 & 52.6 & 54.1 & 46.4 & 43.1 & 44.7\\ 
 Human  & \textbf{89.7} & \textbf{86.3} & \textbf{88.0} & \textbf{71.2} & \textbf{75.8} & \textbf{73.4} \\

\bottomrule
\end{tabular}
 \caption{Human performance (\%).}
\label{table:human performance}
\vspace{-0.5em}
\end{table}

\smallskip
\noindent
\textbf{Human Performance.}
To assess human performance on the task of document-level RE on DocRED, we randomly sample $100$ documents from the test set and ask additional crowd-workers to identify relation instances and supporting evidence. Relation instances identified in the same way as Sec.~\ref{Sec: distant} are recommended to the crowd-workers to assist them. 
The original annotation results collected in Sec.~\ref{Sec:Human-Annotated Data Collection} are used as ground truth. We also propose another subtask of jointly identifying relation instances and supporting evidence, and also design a pipeline model.
Table~\ref{table:human performance} shows the performance of RE model and human. Humans achieve competitive results on both the document-level RE task (RE) and the jointly identifying relation and supporting evidence task (RE+Sup), indicating both the ceiling performance on DocRED and the inter-annotator agreement are relatively high. In addition, the overall performance of RE models is significantly lower than human performance, which indicates document-level RE is a challenging task, and suggests ample opportunity for improvement.

\smallskip
\noindent
\textbf{Performance v.s. Supporting Evidence Types.}
Document-level RE requires synthesizing information from multiple supporting sentences. To investigate the difficulty of synthesizing information from different types of supporting evidence, we devide the $12,332$ relation instances in development set into three disjoint subsets: (1) $6,115$ relation instances with only one supporting sentence (denoted as \textit{single}); (2) $1,062$ relation instances with multiple supporting sentences and the entity pair co-occur in at least one supporting sentence (denoted as \textit{mix}); (3) $4,668$ relation instances with multiple supporting sentences and the entity pair do not co-occur in any supporting sentence, which means they can only be extracted from multiple supporting sentences (denoted as \textit{multiple}).
It should be noted that when a model predicts a wrong relation, we do not know which sentences have been used as supporting evidence, thus the predicted relation instance cannot be classified into the aforementioned subsets and computing precision is infeasible. Therefore, we only report recall of the RE model for each subset, which is $51.1\%$ for \textit{single}, $49.4\%$ for \textit{mix}, and $46.6\%$ for \textit{multiple}. This indicates that while multiple supporting sentences in \textit{mix} may provide complementary information, it is challenging to effectively synthesize the rich global information. Moreover, the poor performance on \textit{multiple} suggests that RE models still struggle in extracting inter-sentence relations.

\smallskip
\noindent
\textbf{Feature Ablations.}
We conduct feature ablation studies on the BiLSTM model to investigate the contribution of different features in document-level RE, including entity types, coreference information, and the relative distance between entities (Eq.~\ref{eqn:distance}). Table~\ref{tab:ablation} shows that the aforementioned features all have a contribution to the performance. Specifically, entity types contribute most due to their constraint on viable relation types. Coreference information and the relative distance between entities are also important for synthesizing information from multiple named entity mentions. This indicates that it is important for RE systems to leverage rich information at document level.


\begin{table}
\begin{center}
\small
\begin{tabular}{lcccc}
\toprule
Setting & Ign\,F1 &  Ign\,AUC & F1 & AUC \\
\midrule
BiLSTM & \textbf{48.87} & \textbf{47.61} & \textbf{50.94} & \textbf{50.26}  \\
\midrule
- entity type & 46.81& 44.46&48.70& 47.29\\
- coreference &  47.22 & 44.72 & 49.37& 47.49 \\
- distance & 47.94&  45.57& 50.19 & 48.43 \\
- all features & 44.08 & 39.94 & 46.52 & 43.18 \\
\bottomrule
\end{tabular}
\end{center}
\caption{Feature ablations on dev set (\%).}
\vspace{-0.5em}
\label{tab:ablation}
\end{table}

\smallskip
\noindent
\textbf{Supporting Evidence Prediction.}
We propose a new task to predict the supporting evidence for relation instances. On the one hand, jointly predicting the evidence provides better explainability. On the other hand, identifying supporting evidence and reasoning relational facts from text are naturally dual tasks with potential mutual enhancement. We design two supporting evidence prediction methods: (1) Heuristic predictor. We implement a simple heuristic-based model that considers all sentences containing the head or tail entity as supporting evidence. (2) Neural predictor. We also design a neural supporting evidence predictor. Given an entity pair and a predicted relation, sentences are first transformed into input representations by the concatenation of word embeddings and position embeddings, and then fed into a BiLSTM encoder for contextual representations. Inspired by \citet{DBLP:conf/emnlp/Yang0ZBCSM18}, we concatenate the output of the BiLSTM at the first and last positions with a trainable relation embedding to obtain a sentence's representation, which is used to predict whether the sentence is adopted as supporting evidence for the given relation instance. As Table~\ref{tab:7} shows, the neural predictor significantly outperforms heuristic-based baseline in predicting supporting evidence, which indicates the potential of RE models in joint relation and supporting evidence prediction. 

\begin{table}[htbp]
\begin{center}
\small
\begin{tabular}{lcc}
\toprule
Method & Dev & Test \\
\midrule
Heuristic predictor & 36.21 & 36.76\\
Neural predictor & \textbf{44.07} & \textbf{43.83}\\
\bottomrule
\end{tabular}
\end{center}
\caption{Performance of joint relation and supporting evidence prediction in F1 measurement (\%).}
\label{tab:7}
\end{table}

\smallskip
\noindent
\textbf{Discussion.}
We can conclude from the above experimental results and analysis that document-level RE is more challenging than sentence-level RE and intensive efforts are needed to close the gap between the performance of RE models and human. We believe the following research directions are worth following: (1) Exploring models explicitly considering reasoning; (2) Designing more expressive model architectures for collecting and synthesizing inter-sentence information; (3) Leveraging distantly supervised data to improve the performance of document-level RE.

\section{Related Work}

A variety of datasets have been constructed for RE in recent years, which have greatly promoted the development of RE systems. \newcite{hendrickx2009semeval}, \newcite{doddington2004automatic} and \newcite{walker2006ace} build human-annotated RE datasets with relatively limited relation types and instances. \newcite{DBLP:conf/pkdd/RiedelYM10} automatically construct RE dataset by aligning plain text to KB via distant supervision, which suffers from wrong labeling problem. \newcite{zhang2017position} and \newcite{han2018fewrel} further combine external recommendations with human annotation to build large-scale high-quality datasets. However, these RE datasets limit relations to single sentences.

As documents provide richer information than sentences, moving research from sentence level to document level is a popular trend for many areas, including document-level event extraction~\cite{walker2006ace,DBLP:conf/tac/MitamuraLH15,DBLP:conf/tac/MitamuraLH17}, fact extraction and verification~\cite{DBLP:conf/naacl/ThorneVCM18}, reading comprehension~\cite{DBLP:conf/nips/NguyenRSGTMD16, joshi2017triviaqa,lai2017race}, sentiment classification~\cite{DBLP:conf/acl/PangL04,DBLP:conf/acl/PrettenhoferS10}, summarization~\cite{DBLP:conf/conll/NallapatiZSGX16} and machine translation~\cite{DBLP:conf/emnlp/ZhangLSZXZL18}. 
Recently, some document-level RE datasets have also been constructed. However, these datasets are either constructed via distant supervision~\cite{quirk2017distant,peng2017nary} with inevitable wrong labeling problem, or limited in specific domain~\cite{li2016biocreative,peng2017nary}. In contrast, DocRED is constructed by crowd-workers with rich information, and is not limited in any specific domain, which makes it suitable to train and evaluate general-purpose document-level RE systems.

\section{Conclusion}
To promote RE systems from sentence level to document level, we present DocRED, a large-scale document-level RE dataset that features the data size, the requirement for reading and reasoning over multiple sentences, and the distantly supervised data offered for  facilitating the development of weakly supervised document-level RE. Experiments show that human performance is significantly higher than RE baseline models, which suggests ample opportunity for future improvement.

\section{Acknowledgement}
This work is supported by the National Key Research and Development Program of China (No. 2018YFB1004503), the National Natural Science Foundation of China (NSFC No. 61572273) and China Association for Science and Technology (2016QNRC001). This work is also supported by the Pattern Recognition Center, WeChat AI, Tencent Inc. Han is also supported by 2018 Tencent Rhino-Bird Elite Training Program.

\bibliography{acl2019}
\bibliographystyle{acl_natbib}
\clearpage
\appendix

\section{Appendices}
\label{sec:appendix}

\subsection{Experimental Details}

In this section, we provide more details of our experiments. To fairly compare the results of different models, we optimized all baselines using Adam, with learning rate of $0.001$, $\beta_1=0.9$, $\beta_1=0.999$.  The other experimental hyper-parameters used in our experiments are shown in Table~\ref{table:hyper-parameter}. Additionally, due to the document-level distance between entities, distances are first divided into several bins $\{1,2,..,2^k\}$, where each bin is associated with a trainable distance embedding. 

\begin{table}[thp!]
\centering
\small
\begin{tabular}{l | c}\toprule
Batch size & 40 \\
CNN hidden size & 200 \\
CNN window size & 3 \\
CNN dropout rate  & 0.5\\
LSTM hidden size & 128 \\
LSTM dropout rate  & 0.2\\
Word embedding dimension & 100\\
Entity type embedding dimension & 20 \\
Coreference embedding dimension & 20 \\
Distance embedding dimension & 20  \\
\bottomrule
\end{tabular}
 \caption{Hyper-parameter settings.}
\label{table:hyper-parameter}
\end{table}

\subsection{Types of Named Entities}

In this paper, we adapt the existing types of named entities used in~\newcite{sang2003introduction} to better serve DocRED. These types include ``Person (PER)'', ``Organization (ORG)'', ``Location (LOC)'', ``Time (TIME)'', ``Number (NUM)'', and ``other types (MISC)''. The types of named entities in DocRED and their covered contents are shown in Table \ref{table:entity types}. 

\begin{table}[thp!]
\centering
\small

\begin{tabular}{c | p{0.35\textwidth}}\toprule

Types & Content\\

\midrule

 PER  &  People, including fictional\\ 
 \hline
 \multirow{2}{*}{ORG}  &  Companies, universities, institutions, political or religious groups, etc. \\
  \hline

   & Geographically defined locations, including mountains, waters, etc. \\
 \multirow{3}{*}{LOC} & Politically defined locations, including countries, cities, states, streets, etc. \\
  & Facilities, including buildings, museums, stadiums, hospitals, factories, airports, etc.\\ 
  \hline
 TIME  &  Absolute or relative dates or periods.\\
 \hline
  NUM  &  Percents, money, quantities\\ 
    \hline

   &  Products, including vehicles, weapons, etc.\\
 \multirow{3}{*}{MISC} & Events, including elections, battles, sporting events, etc.\\
 & Laws, cases, languages, etc \\ 

\bottomrule
\end{tabular}
 \caption{Types of named entities in DocRED.}
\label{table:entity types}
\end{table}

\subsection{List of Relations}
We provide the list of relations in DocRED, including Wikidata IDs, relation names and descriptions from Wikidata in Table  \ref{table:relation list1} and
\ref{table:relation list2}.

\begin{table*}[htp]
\centering
\small
\scalebox{0.89}{

\begin{tabular}{p{0.1\textwidth} | p{0.15\textwidth} | p{0.76\textwidth}}\toprule

Wikidata ID & Name & Description\\

\midrule

P6 & head of government & head of the executive power of this town, city, municipality, state, country, or other governmental body \\
P17 & country & sovereign state of this item; don't use on humans \\
P19 & place of birth & most specific known (e.g. city instead of country, or hospital instead of city) birth location of a person, animal or fictional character \\
P20 & place of death & most specific known (e.g. city instead of country, or hospital instead of city) death location of a person, animal or fictional character \\
P22 & father & male parent of the subject. For stepfather, use "stepparent" (P3448) \\
P25 & mother & female parent of the subject. For stepmother, use "stepparent" (P3448) \\
P26 & spouse & the subject has the object as their spouse (husband, wife, partner, etc.). Use "unmarried partner" (P451) for non-married companions \\
P27 & country of citizenship & the object is a country that recognizes the subject as its citizen \\
P30 & continent & continent of which the subject is a part \\
P31 & instance of & that class of which this subject is a particular example and member. (Subject typically an individual member with Proper Name label.) Different from P279 (subclass of) \\
P35 & head of state & official with the highest formal authority in a country/state \\
P36 & capital & primary city of a country, state or other type of administrative territorial entity \\
P37 & official language & language designated as official by this item \\
P39 & position held & subject currently or formerly holds the object position or public office \\
P40 & child & subject has the object in their family as their offspring son or daughter (independently of their age) \\
P50 & author & main creator(s) of a written work (use on works, not humans) \\
P54 & member of sports team & sports teams or clubs that the subject currently represents or formerly represented \\
P57 & director & director(s) of this motion picture, TV-series, stageplay, video game or similar \\
P58 & screenwriter & author(s) of the screenplay or script for this work \\
P69 & educated at & educational institution attended by the subject \\
P86 & composer & person(s) who wrote the music; also use P676 for lyricist \\
P102 & member of political party & the political party of which this politician is or has been a member \\
P108 & employer & person or organization for which the subject works or worked \\
P112 & founded by & founder or co-founder of this organization, religion or place \\
P118 & league & league in which team or player plays or has played in \\
P123 & publisher & organization or person responsible for publishing books, periodicals, games or software \\
P127 & owned by & owner of the subject \\
P131 & located in the administrative territorial entity & the item is located on the territory of the following administrative entity. Use P276 (location) for specifying the location of non-administrative places and for items about events \\
P136 & genre & a creative work's genre or an artist's field of work (P101). Use main subject (P921) to relate creative works to their topic \\
P137 & operator & person or organization that operates the equipment, facility, or service; use country for diplomatic missions \\
P140 & religion & religion of a person, organization or religious building, or associated with this subject \\
P150 & contains administrative territorial entity & (list of) direct subdivisions of an administrative territorial entity \\
P155 & follows & immediately prior item in some series of which the subject is part. Use P1365 (replaces) if the preceding item was replaced, e.g. political offices, states and there is no identity between precedent and following geographic unit \\
P156 & followed by & immediately following item in some series of which the subject is part. Use P1366 (replaced by) if the item is replaced, e.g. political offices, states \\
P159 & headquarters location & specific location where an organization's headquarters is or has been situated \\
P161 & cast member & actor performing live for a camera or audience [use "character role" (P453) and/or "name of the character role" (P4633) as qualifiers] [use "voice actor" (P725) for voice-only role] \\
P162 & producer & producer(s) of this film or music work (film: not executive producers, associate producers, etc.) [use P272 to refer to the production company] \\
P166 & award received & award or recognition received by a person, organisation or creative work \\
P170 & creator & maker of a creative work or other object (where no more specific property exists) \\
P171 & parent taxon & closest parent taxon of the taxon in question \\
P172 & ethnic group & subject's ethnicity (consensus is that a VERY high standard of proof is needed for this field to be used. In general this means 1) the subject claims it him/herself, or 2) it is widely agreed on by scholars, or 3) is fictional and portrayed as such). \\
P175 & performer & performer involved in the performance or the recording of a work \\
P176 & manufacturer & manufacturer or producer of this product \\
P178 & developer & organisation or person that developed this item \\
P179 & series & subject is part of a series, whose sum constitutes the object \\
P190 & sister city & twin towns, sister cities, twinned municipalities and other localities that have a partnership or cooperative agreement, either legally or informally acknowledged by their governments \\
P194 & legislative body & legislative body governing this entity; political institution with elected representatives, such as a parliament/legislature or council \\
P205 & basin country & country that have drainage to/from or border the body of water \\

\bottomrule

\end{tabular}
}
 \caption{Relation list (\Rmnum{1}), including Wikidata IDs, names and descriptions of relations in DocRED.}
 \label{table:relation list1}

\end{table*}

\begin{table*}[thp!]
\centering
\small
\scalebox{0.89}{

\begin{tabular}{p{0.1\textwidth} | p{0.15\textwidth} | p{0.76\textwidth}}\toprule

Wikidata ID & Name & Description\\

\midrule
P206 & located in or next to body of water & sea, lake or river \\
P241 & military branch & branch to which this military unit, award, office, or person belongs, e.g. Royal Navy \\
P264 & record label & brand and trademark associated with the marketing of subject music recordings and music videos \\
P272 & production company & company that produced this film, audio or performing arts work \\
P276 & location & location of the item, physical object or event is within. In case of an administrative entity use P131. In case of a distinct terrain feature use P706. \\
P279 & subclass of & all instances of these items are instances of those items; this item is a class (subset) of that item. Not to be confused with P31 (instance of) \\
P355 & subsidiary & subsidiary of a company or organization, opposite of parent company (P749) \\
P361 & part of & object of which the subject is a part. Inverse property of "has part" (P527). See also "has parts of the class" (P2670). \\
P364 & original language of work & language in which a film or a performance work was originally created. Deprecated for written works; use P407 ("language of work or name") instead. \\
P400 & platform & platform for which a work has been developed or released / specific platform version of a software developed \\
P403 & mouth of the watercourse & the body of water to which the watercourse drains \\
P449 & original network & network(s) the radio or television show was originally aired on,  including \\
P463 & member of & organization or club to which the subject belongs. Do not use for membership in ethnic or social groups, nor for holding a position such as a member of parliament (use P39 for that). \\
P488 & chairperson & presiding member of an organization, group or body \\
P495 & country of origin & country of origin of the creative work or subject item \\
P527 & has part & part of this subject. Inverse property of "part of" (P361). \\
P551 & residence & the place where the person is, or has been, resident \\
P569 & date of birth & date on which the subject was born \\
P570 & date of death & date on which the subject died \\
P571 & inception & date or point in time when the organization/subject was founded/created \\
P576 & dissolved, abolished or demolished & date or point in time on which an organisation was dissolved/disappeared or a building demolished; see also discontinued date (P2669) \\
P577 & publication date & date or point in time a work is first published or released \\
P580 & start time & indicates the time an item begins to exist or a statement starts being valid \\
P582 & end time & indicates the time an item ceases to exist or a statement stops being valid \\
P585 & point in time & time and date something took place, existed or a statement was true \\
P607 & conflict & battles, wars or other military engagements in which the person or item participated \\
P674 & characters & characters which appear in this item (like plays, operas, operettas, books, comics, films, TV series, video games) \\
P676 & lyrics by & author of song lyrics; also use P86 for music composer \\
P706 & located on terrain feature & located on the specified landform. Should not be used when the value is only political/administrative (provinces, states, countries, etc.). Use P131 for administrative entity. \\
P710 & participant & person, group of people or organization (object) that actively takes/took part in the event (subject).  Preferably qualify with "object has role" (P3831). Use P1923 for team participants. \\
P737 & influenced by & this person, idea, etc. is informed by that other person, idea, etc., e.g. "Heidegger was influenced by Aristotle". \\
P740 & location of formation & location where a group or organization was formed \\
P749 & parent organization & parent organization of an organisation, opposite of subsidiaries (P355) \\
P800 & notable work & notable scientific, artistic or literary work, or other work of significance among subject's works \\
P807 & separated from & subject was founded or started by separating from identified object \\
P840 & narrative location & the narrative of the work is set in this location \\
P937 & work location & location where persons were active \\
P1001 & applies to jurisdiction & the item (an institution, law, public office ...) belongs to or has power over or applies to the value (a territorial jurisdiction: a country, state, municipality, ...) \\
P1056 & product or material produced & material or product produced by a government agency, business, industry, facility, or process \\
P1198 & unemployment rate & portion of a workforce population that is not employed \\
P1336 & territory claimed by & administrative divisions that claim control of a given area \\
P1344 & participant of & event a person or an organization was a participant in, inverse of P710 or P1923 \\
P1365 & replaces & person or item replaced. Use P1398 (structure replaces) for structures. Use P155 (follows) if the previous item was not replaced or if predecessor and successor are identical. \\
P1366 & replaced by & person or item which replaces another. Use P156 (followed by) if the item is not replaced (e.g. books in a series), nor identical \\
P1376 & capital of & country, state, department, canton or other administrative division of which the municipality is the governmental seat \\
P1412 & languages spoken, written or signed & language(s) that a person speaks or writes, including the native language(s) \\
P1441 & present in work & work in which this fictional entity (Q14897293) or historical person is present \\
P3373 & sibling & the subject has the object as their sibling (brother, sister, etc.). Use "relative" (P1038) for siblings-in-law (brother-in-law, sister-in-law, etc.) and step-siblings (step-brothers, step-sisters, etc.) \\

\bottomrule
\end{tabular}}
 \caption{Relation list (\Rmnum{2}), including Wikidata IDs, names and descriptions of relations in DocRED.}
\label{table:relation list2}
\end{table*}

\end{document}


\section*{Supplemental Material}
\label{sec:supplemental}

\section{Named Entity Recognition}
Recognizing all named entity mentions in the document is a core sub-task of document-level RE. The following state-of-the-art models are evaluated on the human-annotated data in DocRED on NER task: (1) BiLSTM-CRF ~\cite{huang2015bidirectional} which encodes sentences with BiLSTM and performs sequence labeling with a CRF layer; (2) Char-BiLSTM-CRF~\cite{ma2016end} which further utilizes CNN to compose character embeddings into word-level representations; (3) \BERTBASE and \BERTLARGE~\cite{devlin2018bert} fine-tuned on the human-annotated data in DocRED.
We find there is still ample opportunity for improvement for NER task on our dataset.

\begin{table}
\begin{center}
\small
\begin{tabular}{lc}
\toprule
Model &  F1\\
\midrule
BiLSTM-CRF & 86.16\\
Char-BiLSTM-CRF & 87.04\\
\BERTBASE  & 90.21\\
\BERTLARGE  &  \textbf{90.50} \\
\bottomrule
\end{tabular}
\end{center}
\caption{NER performance on DocRED. }
\label{Table:NER result}
\end{table}


\bibliography{acl2019}
\bibliographystyle{acl_natbib}